# Deep Learning Classification of Photoplethysmogram Signal for Hypertension Levels

**Nida Nasir ¹, Mustafa Sameer ², Feras Barneih ³, Omar Alshaltone ³, Muneeb Ahmed ⁴**

¹  Universite Gustave Eiffel (IFSTTAR), Lille, France
²  National Institute of Technology, Patna, India.
³  Research Institute of Science and Engineering, University of Sharjah, United Arab Emirates.
⁴  Jamia Millia Islamia University, New Delhi, Delhi, India
*  Correspondence: nida.nasir@univ-eiffel.fr

**Abstract:** Continuous photoplethysmography (PPG)-based blood pressure monitoring is necessary for healthcare and fitness applications. In Artificial Intelligence (AI), signal classification levels with the machine and deep learning arrangements need to be explored further. Techniques based on time-frequency spectra, such as Short-time Fourier Transform (STFT), have been used to address the challenges of motion artifact correction. Therefore, the proposed study works with PPG signals of more than 200 patients (650+ signal samples) with hypertension, using STFT with various Neural Networks (Convolution Neural Network (CNN), Long Short-Term Memory (LSTM), Bidirectional Long Short-Term Memory (Bi-LSTM), followed by machine learning classifiers, such as, Support Vector Machine (SVM) and Random Forest (RF). The classification has been done for two categories: Prehypertension (normal levels) and Hypertension (includes Stage I and Stage II). Various performance metrics have been obtained with two batch sizes of 3 and 16 for the fusion of the neural networks. With precision and specificity of 100% and recall of 82.1%, the LSTM model provides the best results among all combinations of Neural Networks. However, the maximum accuracy of 71.9% is achieved by the LSTM-CNN model. Further stacked Ensemble method has been used to achieve 100% accuracy for Meta-LSTM-RF, Meta- LSTM-CNN-RF and Meta- STFT-CNN-SVM.

## 1. Introduction

For many years, photoplethysmography (PPG) has been used to determine how much light is reflected or absorbed in tissues by blood vessels. Therefore, the PPG provides inexpensive and flexible blood pressure measurements [1]. Moreover, identification of blood oxygen saturation, BP estimation, heart rate, respiration, cardiac output, arterial aging, endothelial controller, autonomic function, and microvascular blood flow are some of the elements of cardiovascular surveillance that might be expanded [2], [3]. Many PPG signals have been discovered, and they have been linked to aging and cardiovascular diseases [4]. PPG signals are captured from the different parts of the body, such as the finger, earlobe, forehead, toes, etc. [6]. The PPG sensor works with veins, arteries, and several capillaries for blood pressure measurements. Moreover, deep learning is the most recent success of the machine learning era. Initially, it presented near-human competencies, but nowadays, it can extend to super-human abilities in many scenarios, such as object detection [7], voice-to-text translations [8], anomaly detection [9], recognition, emotion recognition from audio or video recordings [10], etc. Remote services such as telemedicine, telehealth, and telecare improve healthcare, especially for the elderly, by continuously monitoring and access to real-time care. Moreover, sensors installed in homes offer a great opportunity for monitoring activities and behaviors. This information could be sent to family members or care services during an emergency [11]. However, typical machine learning algorithms are insufficient for processing extensive healthcare data to detect problems. Researchers presented a modern healthcare monitoring framework based on the cloud environment and a big data analytics engine to properly store and analyze healthcare data and increase classification accuracy [13]. However, existing hypertension risk scoring models do not consider occupational factors; hence they cannot be used to calculate the risk of hypertension in steelworkers. Therefore, a risk score algorithm for hypertension in steelworkers must be developed. Thus, this may be considered another important use for this sector [14].

## 2. Background

The recent development and progress in this field have made it a potential candidate for advanced healthcare. One of the significant problems of estimating PPG is the need for regular calibration related to accuracy loss in existing methods for BP estimation from PPG [15]. For the time-series BP data [16], a Recurrent Neural Network (RNN) through Long Short-Term Memory (LSTM) has been utilized in building the prototype. PPG and Electrocardiogram (ECG) were considered in the role of inputs; moreover, Pulse Transit Time (PTT) with few different characteristics was utilized as factors for estimating blood pressure. When compared to other current approaches, this strategy improved BP prediction. Another study examined the potential of employing raw PPG data to identify arrhythmia with good results, demonstrating the feasibility of using raw PPG signals as deep learner inputs [17]. In [18], the authors developed a new spectro-temporal deep neural network that has used the PPG signal and its first and second derivatives as inputs. The residual connections in the neural network model resulted in a Mean Absolute Error (MAE) of 6.88 and 9.43 for Diastolic BP and Systolic BP.

For BP estimation using LSTM, various studies have been used, such as a hierarchical Artificial Neural Network – Long Short-Term Memory (ANN-LSTM) model based on waveforms [19], the real-time system using LSTM to monitor the quality of PPG signals [20], etc. Moreover, LSTM may retain valuable signals by filtering out noise signals when data-driven; therefore, it is suggested to train a two-layer LSTM model to predict pulse signals. To compute the heart rate, the authors proposed to use several synthetic signals created using the algorithm they designed to pre-train the model and pure periodic signals filtered from LSTM. Experiential findings on a public database demonstrate the efficacy of their suggested technique, which may be used as a reference for heart rate estimates [21], [22].

For accurate PPG cardiac period segmentation, a deep RNN based on Bidirectional Long Short-Term Memory (BiLSTM) was described in [23] [24], which was the first time a deep learning-based method was used for Pulse Rate Variability (PRV) extraction from significantly weak and noisy PPG signals. Three key PRV indicators, i.e., peak intervals, pulse intervals, and Instantaneous Heart Rates (IHR), were computed. The suggested algorithm's performance in estimating and recovering PRV indices is promising when associated with state-of-the-art techniques on a dataset of 48 patients. Convolutional Neural Networks (CNN) and Bi-LSTM models obtained average classification accuracy of 93.9 percent and 97.2 percent, respectively [16,17]. In various vision-based applications, CNN has demonstrated promising results [27].

Researchers used a green LED PPG sensor to record physiological data in both static and activity modes [28]. As a result, they used a 5-minute short-term assessment. These signals were separated into five pieces, each lasting one minute. Researchers used time-domain analytic criteria to compute heartbeats per minute and Heart Rate Variability (HRV) [29]. In frequency domain analysis, the related technique of Short-Time Fourier Transform (STFT) coupled with Power Spectral Density (PSD) is used to identify HRV [28]. Moreover, other researchers suggested an STFT hardware realization approach to enhance epilepsy seizure detection. Using STFT output, they managed to extract the features of the signal and then feed it into CNN and Bi-LSTM to detect seizures. During the comparison of STFT output obtained by suggested hardware design vs. output generated via simulation, a maximum inaccuracy of 0.13 percent was found [25].

In Table 1, a review of several studies has been summarised. Most studies have used handcraft feature extraction techniques followed by machine learning classifiers. The major drawback of these techniques is that they require specialists to select optimal features. Deep learning methodologies overcome this issue. At the same time, the major disadvantage of deep learning techniques is their long training time. This study proposed deep learning-based feature extraction models having minimal complexity to overcome these issues. The main methodology of this study is indicated in Fig. 1.

Main Contributions of the proposed study:

(i) Features have been extracted using deep learning architectures followed by two robust machine learning classifiers, i.e., RF and SVM.

(ii) The authors have compared different deep learning techniques such as Convolution Neural Network, Short-Time Fourier Transform, Long Short-Term Memory, and Bidirectional Long Short-Term Memory.

(iii)The models have been designed using the least trainable parameters, making them useful for real-time implementation.

(iv) Prehypertension and Hypertension are the two categories for which the classification has been made.

(v) Stacked Ensemble method has been used to improve performance metrics.

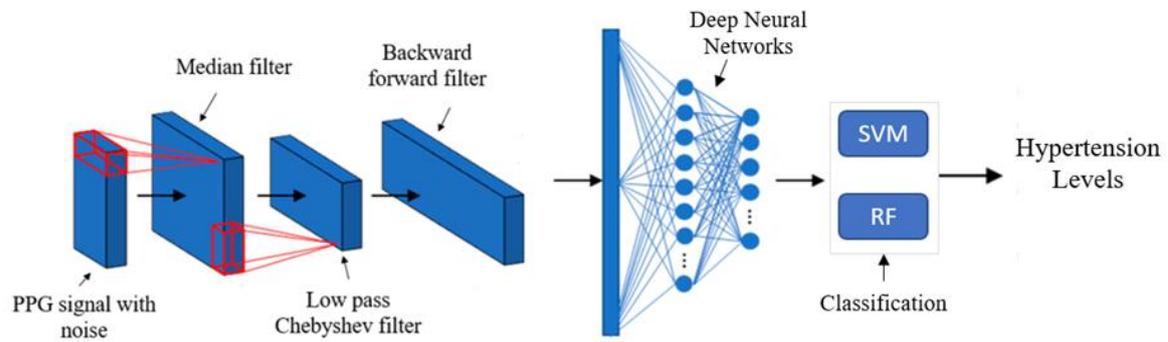

**Figure 1.** Methodology of this study

Many recent publications have studied the comparison of various classifiers across an extensive collection of medical application datasets. In [30], the authors evaluated more than 170 classifiers, and the result shows that RF can give the best results, followed by SVM. Moreover, in [31], a detailed comparison of RF and SVM with other studies showed their performance in medical classification problems. Hence, based on previous detailed studies, SVM and RF have been chosen as machine learning classifiers for this study. A single ML classifier would have worked for correlation with NN but using more than one classifier gives this study an edge.

The paper is organized as follows. All the components of the methodology of this work are stated in Section II. The performance of the proposed methodology is investigated in Section III. Section IV discusses the pros and cons and the study's future scope. Finally, Section V concludes the paper.

**Table 1.** Summary and Review of Several Studies.

| Ref | Methodology | Dataset | Datatype | No of patients/samples | Attributes/Findings |
|-----|-------------|---------|----------|------------------------|---------------------|
| [32] | Deep learning (Spectro-temporal ResNet) | MIMIC III [33] | PPG | 510 Subjects | Achieved MAE of 9.43 and 6.88 for systolic and diastolic blood pressure respectively. |
| [34] | PPG and FFT-based neural networks | MIMIC II [35] | PPG | (69 subjects from MIMC II) plus 23 volunteers | This approach is quick and reliable, and it has the potential to be utilized for pulse wave analysis in addition to blood pressure determination. |
| [36] | PTT approach, classical ML (AdaBoost) | MIMIC II [35] | PPG | (1000 subjects) | An algorithm for the cuff-less blood pressure estimation can potentially enable mobile healthcare gadgets to monitor the BP continuously. The suggested technique used the criteria of the Association for the Advancement of Medical Instrumentation (AAMI) and the British Hypertension Society (BHS). According to AAMI standards, DBP and MAP values are estimated using this technique. |
| [37] | RNN, LSTM | Proprietary data | PPG | 84 subjects | A novel deep RNN along with bidirectional LSTM. Achieved RMSE of 3.90 and 2.66 mmHg for systolic and diastolic blood pressure, respectively. |
| [38] | MTGPs and ANNs | MIMIC [39] | PPG | 100 subjects | It proposed a new model using MTGP that gave better results than other neural network models. |

| Ref | Methodology | Dataset | Datatype | No of patients/samples | Attributes/Findings |
|---|---|---|---|---|---|
| [40] | CNN, RNN | MIMIC [41] | PPG | 90 ICU patients | End-to-end neural network model consisting of CNN and RNN. The proposed method is simple and achieves better accuracy. According to the results, the SBP RMSE is 4.643 mmHg, whereas the DBP RMSE is just 3.307 mmHg. |
| [42] | CNN | MIMIC III [33] | PPG, ECG | 22 subjects | Used both ECG and PPG to estimate BP. Showed that CNN could be trained simultaneously with several patient's data. The results indicated that the average root mean square error (RMSE) for SBP and DBP was 5.42 mmHg and 7.81 mmHg, respectively. |
| [19] | ANN-LSTM | MIMIC [39] | PPG, ECG | 39 ICU patients | Used ECG and PPG waveforms to estimate BP. Their model extracted features without the need for any feature engineering while performing better than conventional methods. There is an average error of 1.10 mmHg for SBP estimate and an error of 0.58 and 0.85 for DBP estimation. |
| [43] | Logistic Regression, AdaBoost Tree, Bagged Tree, and KNN | MIMIC [39] | PPG, ECG | 121 subjects | The F1 scores for classification of normotension vs. pre-hypertension were 84.34 percent, 94.84 percent for normotension vs. hypertension, and 88.49 percent for normotension plus pre-hypertension vs. hypertension. |
| [44] | Wavelet transform, SVM | MIMIC II [35] | PPG | 20 normal subjects and 6 CAD | The classification results for MIMIC-II are (Accuracy=89%, Sensitivity=86%, Specificity=90%), and for the Local hospital is (Accuracy=93%, Sensitivity=92%, Specificity=94%) |
| [45] | Kalman filter | MIMIC II [35] | PPG | 21 subjects | The suggested model outperformed the competition in terms of measurement accuracy. Wearable sensor systems can benefit from this low-complexity method because of its low computational complexity. |
| [46] | Autoregression (AR) | 1-MIMIC 2-BIDMC 3-CapnoBase 4-Local dataset | PPG | 90 patients, 53 critically ill patients, 42 individuals, 30 healthy individuals | Introduced an autoregression model for wearable devices using four different PPG datasets. Their AR model can recreate the PPG signal's pattern behavior, trend, and statistical features according to the results. The produced PPG samples also show comparable periodic autocorrelation peaks to the original ones, even with a 14% error rate. |
| [47] | Deep learning | MIMIC III [33] | PPG | 140 subjects | The simulation results show that the suggested method reduces AAE by 1.07 and standard deviation by 1.87. |
| [48] | Bidirectional Recurrent Denoising Auto-Encoder (BRDAE) | 1- MIMIC III 2-ECG and PPG Measurement Over 24 Hours | PPG | 1- 250 subjects  2- 9 healthy males | It offers benefits beyond standard denoising by allowing PPG feature accentuation in pulse waveform analysis. In terms of correlation to reference and root-mean-squared error, the denoised PPG performed significantly better heart rate detection than the original PPG. |

| Ref | Methodology | Dataset | Datatype | No of patients/samples | Attributes/Findings |
|-----|-------------|---------|----------|------------------------|---------------------|
| [49] | Deep Learning (LSTM) | MIMIC-II [35] | PPG | 20 subjects | Two different standard hospital datasets revealed absolute errors mean and absolute error standard deviation for systolic and diastolic blood pressure measurements, which outperformed previous studies by the uses of three layers that done by deep neural network to evaluate BP using PPG signals. |
| [50] | CWT, CNN | MIMIC-III [33] | PPG | N/A | The model's accuracy and robustness were improved by validating the results. Both the British Hypertension Society's (BHS) standard and the American National Standard from AAMI were fulfilled by our model to estimate SBP. |
| [51] | LMA, BRA, and SCGA | Dr. Raymond Lam, Glax-oSmith Kline dataset | PPG | 250 subjects | A novel model to predict SBP using three different training algorithms compared the results with five different machine learning algorithms. The results were verified to improve the model's accuracy and resilience. As for SBP estimate, our model fulfilled both the BHS standard and the AAMI at grade A. |
| [52] | MTM, ANN | MIMIC | PPG | 90 ICU patients | The suggested technique achieves higher accuracy; the mean absolute error for systolic BP is 4.02 2.79 mmHg and 2.27 1.82 mmHg for diastolic BP. |
| [53] | DTC, RF, LDA, and LSVM | PPG Blood Pressure, IEEE Dataport | PPG | 219 subjects | Used four different classifier algorithms and compared each other results to diagnose hypertension disease. Results found that C4.5 decision tree and random forest classifiers achieved the best results. |

**BRA**: Bayesian Regularization Algorithm, **BIDMC**: Beth Israel Deaconess Medical Centre **CWT**: Continuous Wavelet Transform, **DTC**: Decision Tree Classifier, **EMD**: Empirical Mode Decomposition, **FFT**: Fast Fourier Transform, **HWT**: Harr Wavelet Transform, **ICU**: Intensive Care Unit, **KNN**: K-Nearest Neighbours, **LDA**: Linear Discriminant Analysis, **LSVM**: Linear Support Vector Machine, **LMA**: Levenberg-Marquardt Algorithm, **MTGP**: Multi-Task Gaussian Process, **MIMIC**: Medical Information Mart for Intensive Care, **ML**: Machine Learning, **ResNet**: Residual Neural Network, **RMSE**: Root Mean Square Error, **SVD**: Singular Value Decomposition, **SCGA**: Scaled Conjugate Gradient Algorithm, **VCS**: Variance Characterization Series.

## 3. Methodology

This section discusses the dataset, pre-processing steps, and different Neural Network (NN) architectures layers. The methodology of this study is depicted in Fig. 1. PPG signal quality is affected by both physiological factors (such as skin qualities) and environmental factors (such as power line interferences and motion artifacts), which appear as a noise in the signal [54]. This study combined all kinds of hypertension data and analyzed it with data of normal people. Hence, we have two categories: Prehypertension and Hypertension. Signal pre-processing reduces these noisy components, which have detrimental impacts on system performance and the extraction of PPG cycles. For the proposed study, pre-processing includes three filters: Median filter, low pass Chebyshev filter, and backward, forward filter, as demonstrated in Fig. 1. After this, deep learning used on processed data intended for feature extraction and resulting in the hypertension levels by classification. The proposed method implemented using Python deep learning library TensorFlow and Keras in Google Colab. For pre-processing Python's SciPy has been used.

### 3.1. Dataset

The data was gathered at Guilin People's Hospital in Guilin, China, with prior patients' approval and the ethics committee's approval [55]. This dataset combines the complete medical data of 219 patients admitted to the mentioned hospital. It includes data related to many conditions like cardiovascular diseases (CVD), diabetes, cerebral infarction, and insufficient blood supply. The dataset used here is available to access. It intends to assist in predicting ailments from PPG, signal quality assessment, etc.

It is established from 219 patients (ages 20 to 89 years), including 657 PPG segments and their hypertension and diabetes records.  Moreover, this dataset includes all details that can act as a feature in artificial intelligence techniques, for instance, their age, height, weight, systolic and diastolic blood pressure (SBP and DBP), heart rate, and body mass index (BMI). The classification is done by high blood pressure and diabetes level, taken from the same hand. The signal's sample rate is 1 kHz, and all the patient data contains 2100 sampling points, equating to 2.1 seconds of data. In a total of 219 patients, there are 105 males and 114 females. According to the hypertension categorization, 36.5% of patients have typical blood pressure, whereas 38.8%, 15.5% stage 1 hypertension, and stage 2 hypertension, respectively, and 9.2 percent have: pre-hypertension, as shown in Fig. 2. The percentage of the patients that suffer from cerebral infarction effects is 9.5. In contrast, cerebrovascular disease and insufficiency of cerebral blood supply affected only 5% and 6.8% of patients, correspondingly. The non-uniform character of the data is demonstrated by the following distribution, as shown in Fig 2. The training and testing were 70% and 30% respectively.

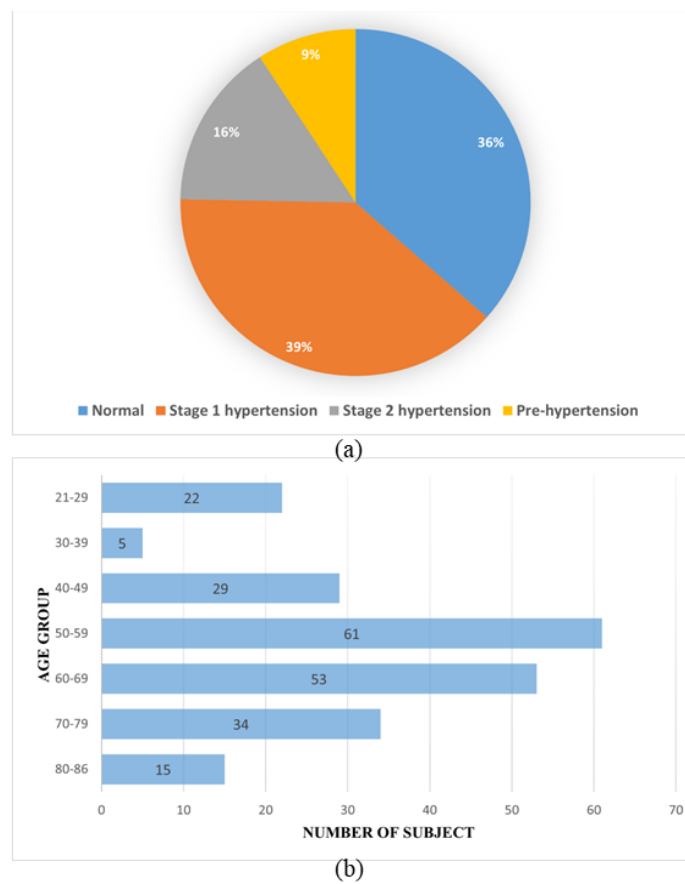

(a)

(b)

**Figure 2.** Distribution of the dataset used based on a) type of hypertension, b) patients age [55].

*3.2. Pre-processing*

PPG signals in the dataset contain noises that influence the quality of the signal, which can lead to errors. Hence, the signal is pre-processed by applying different methods to tackle this problem and to get an optimal result. Denoising, detrending, normalizing, and clipping methods have pre-processed the raw PPG. Afterward, denoising is used to filter the valuable data from high-frequency noise produced throughout signal capture by external or internal variables, e.g., finger movement, skin thickness, etc. [56].  The signal denoising process is passed through a median filter with a kernel size of three. The odd number is preferable for the kernel value to determine the mid-value. Afterward, the median filter's output is sent over to Chebyshev 2. The Chebyshev 2 low pass filter is of order four and has a cut-off frequency of 25Hz with a minimum attenuation needed in the stopband of 10. The sampling frequency (i.e., 1000 Hz) of the Chebyshev 2 filter is also passed as an argument. Finally, the backward-forward filter is used to fix the edges of a signal that has been filtered using the median and Chebyshev filters. Detrending eliminates data characteristics that cause the signal to be distorted [57], [58], [59].

### 3.2.1. Short-Time Fourier Transform

The Short-Time Fourier transform (STFT) is a widely used speech and audio signal analysis, modification, and synthesis technique. Standard Fourier transforms only convert the signal from the time domain to the frequency domain. Thus, the time domain information of the analyzed signal is lost. Therefore, Short-time Fourier transform (STFT) is used for time-frequency analysis as it provides both time and frequency information. It works by slicing the signal into multiple windows and multiplying it by a half cosine function, pads zeros, and computing the Fourier transform to each window. To compute the STFT of any signal, the following window function can be used [60]:

$$X(m,k) = \sum_{n=-\infty}^{\infty} x[n]w(n - Lm)e^{-j\frac{2\pi}{N}kn} \tag{1}$$

Where $X(m,k)$ is the STFT of the input signal $x[n]$ at time $n$ while $w[n]$ is the window function of length N and step size of L.

### 3.2.2. Independent Component Analysis

Independent Component Analysis (ICA) is a multivariate data analysis approach that describes an extensive database of samples. The independent components of the observed data are the variables in the model that are non-Gaussian and mutually independent. These are referred to as sources or factors. When traditional approaches like Principal Component Analysis (PCA) fail, ICA is a robust tool that can uncover the underlying variables or causes. The separate source signals are discovered using ICA from a series of linear mixes of the underlying sources. Assumption of observing $n$ linear mixes of $x_1, x_2… x_n$ independent components, as indicated in the equation below [61].

$$x_j = a_{j1}s_1 + a_{j2}s_2 + ….a_{jn}s_n \tag{2}$$

The independent component $s_i$ and the variable $x_j$ are random variables in this equation, whereas $x_j(t)$ and $s_i(t)$ are samples of random variables. The independent component and the variable are also assumed to have zero mean, simplifying the issue to the model zero-mean, which is provided by the equation below.

$$\hat{X} = X - E(X) \tag{3}$$

Let $x$ and $s$ represent the random vectors $x_1, x_2, x_3 … x_n$ and $s_1, s_2, … s_n$, respectively. Let A be the matrix containing the $a_{ij}$ entries, as shown in the equation below.

$$x = \sum_{i=1}^{n} a_i s_i \tag{4}$$

$$x = As \tag{5}$$

The equation represents independent component analysis (ICA), a method used to compute both the matrix A and independent components when only the measurement variable x is available. The model assumes independent and non-Gaussian components. ICA algorithmically decomposes a multidimensional data vector into statistically independent components. It finds application in eliminating artifacts and decomposing PPG recorded data into distinct component signals from various sources. Data processed by ICA originates from diverse sources such as digital images, document databases, economic indicators, and psychometric measures. Often, data are presented as parallel signals or time series, a challenge known as blind source separation. Examples include mixtures of simultaneous voice sounds captured by multiple microphones, brain waves recorded by different sensors, interfering radio signals received by a mobile phone, or parallel time series obtained from industrial processes [62], [63], [64].

### 3.3. Feature Extraction

After pre-processing, the extracted features are evaluated in the following NNs to classify the BP from the PPG signals.

3.3.1. CNN ArchitectureA CNN architecture comprises a series of separate layers that use a differentiable function to turn the input volume into an output volume (e.g., containing the class scores). There are a few different sorts of layers that are broadly employed. As illustrated in Fig. 3, the kernel with width $k$ is slid in one dimension along the axis of time of length $n$.

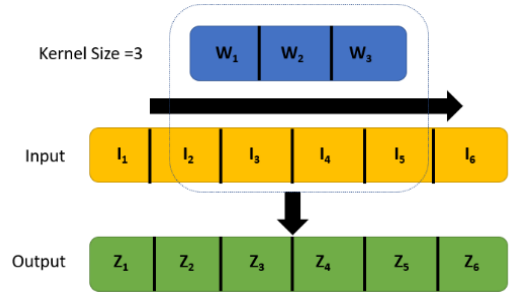

**Figure 3.** Convolution layer Architecture.

A typical CNN architecture consists of convolution and pooling layers alternated with one or more fully linked layers at the end [27]. In some circumstances, a global average pooling layer may substitute a fully connected layer. Diverse regulatory units such as batch normalization and dropout are also added to enhance CNN performance in addition to different mapping functions [65].

Sequence data, such as time series and natural language, have a long-term dependence that necessitates preserving essential information from any state. RNNs are meant to compensate for the shortcomings of conventional neural networks in dealing with situations involving sequential data [25], [66]. Nevertheless, RNN's performance begins to deteriorate as the sequence grows longer. One type of RNN designed to address this problem is the LSTM. The LSTM module, also known as a memory cell, has a particular composition that permits information to endure and pass on over time, as seen in Figure 4. Also, input and output gates along with cell state, are all included in the construction of an LSTM memory cell [67]. Moreover, various layers of connection for LSTM have been presented in Fig. 5(a).

$$i_t = \sigma(W_i.[h_{t-1}, x_t] + b_i) \tag{6}$$

$$f_t = \sigma(W_f.[h_{t-1}, x_t] + b_f) \tag{7}$$

$$o_t = \sigma(W_o.[h_{t-1}, x_t] + b_o) \tag{8}$$

$$c_t = f_t \circ c_{t-1} + i_t \circ \tilde{c}_t \tag{9}$$

$$\tilde{c}_t = \sigma_c(W_c x_t + U_c h_{t-1} + b_c) \tag{10}$$

$$h_t = o_t \circ \sigma_h(c_t) \tag{11}$$

Symbol's definitions are explained in Table 2 below:

**Table 2.** LSTM EQUATIONS SYMBOL DEFINITION.

| Symbol | Definition |
|---|---|
| $i_t$ | input/update gate's activation vector |
| $f_t$ | Forget gate's activation vector |
| $o_t$ | Output gate's activation vector |
| $h_t$ | Hidden state vector |
| $\hat{c}_t$ | Cell input activation vector |
| $c_t$ | Cell state vector |
| $x_t$ | Input vector to LSTM unit |
| $\sigma$ | Sigmoid function |
| $W_x$ | weight matrices and bias vector parameter |

### 3.3.2. BILSTM

Bi-directional LSTM (BiLSTM) has been very useful when required input context. It comes in handy for projects like emotion categorization. Information travels from backward to forward in a unidirectional LSTM. However, BiLSTM uses two hidden states to flow information not just backward to forward but also forward to backward. As a result, BiLSTM has a greater analysis level of the input. BiLSTMs were employed to increase the input data that the network could utilize. RNN with LSTM and RNN with BiLSTM structure have been discussed [24]. The reverse direction states that inputs are unrelated to the outcomes of these two stages. The BiLSTM structure is shown in Fig. 5(b). Input data from the past and future of the current time frame may be utilized using two-time directions. Standard RNN, on the other hand, requires delays to account for future data [68], [69]. As the BiLSTM model collects more information from both directions of the target word, it allows BiLSTM to obtain more semantic information. The BiLSTM model increases the accuracy of evaluating sentiment polarity when compared to the regular LSTM model.

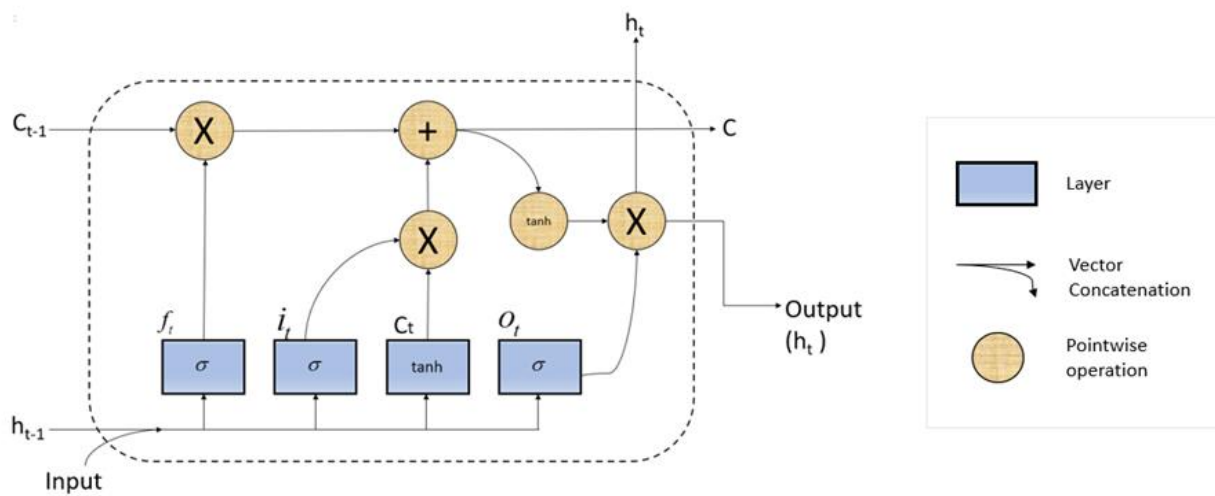

**Figure 4.** The flowchart of LSTM module.

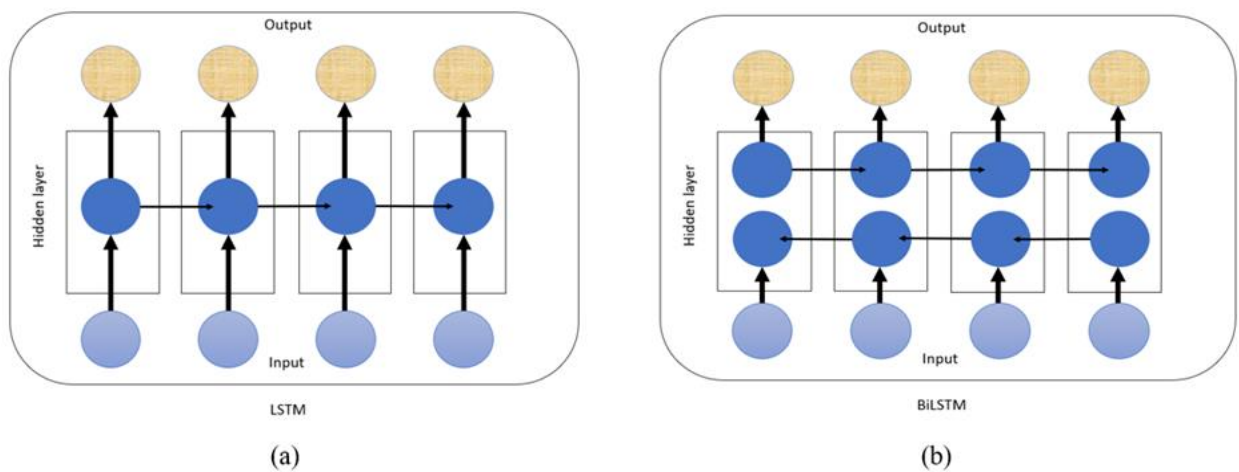

**Figure 5.** Illustration of (a) LSTM, and (b) BiLSTM.

*3.4. Neural Network Layers*

The primary layer in the architecture operates as a convolution layer, employing a specialized linear operation with a sliding kernel for feature extraction. Image data, being two-dimensional, necessitates 2D convolution, while time-series data, being one-dimensional, requires 1D convolution. In this study, 1D convolution is applied to time-series PPG data. The kernel, with width 'k', slides along the time axis of length 'n'. PPG signals are collected via electrodes on the subject's head, with the kernel width corresponding to the number of electrodes. The kernel iterates through the time series, multiplying its values with corresponding time series values to generate new univariate time series values. This process repeats for each block, extracting features for subsequent series. The resulting sum becomes the new value of the univariate time series, as shown in Fig. 3. The kernel then moves to the next block and repeats the process to the end of the block, extracting features for the new series.

$$n_{out} = \left[ \frac{n_{in} + 2p - k}{s} \right] + 1$$

(12)

Where $n_{out}$: number of output features, $n_{in}$: number of input features, p: convolution padding size, s: convolution stride size and k: convolution kernel size [70].

The subsequent layer is the batch normalization layer, responsible for standardizing inputs to the convolutional layer. In our modeling approach, this layer normalized the PPG signal inputs to have a mean of zero and a standard deviation of one. This transformation mitigated internal covariance shift, enhancing the overall speed and efficiency of the training process [71]. Also, it is used to make the output of the previous layers more natural. Learning becomes more efficient when batch normalization is utilized, and it may also be used as a regularization to prevent model overfitting. Finally, a layer is added to the sequential model to standardize the inputs and outputs. It may be applied at various locations within the model's layers. It is frequently placed directly following the definition of the sequential model and before the convolution and pooling layers [72]. Following the convolutional layer, the pooling layer is utilized to reduce redundant features. This layer is typically applied post-convolution. Two commonly employed methods are average pooling and max pooling. Average pooling computes the mean value of the neurons, while max pooling selects the maximum value. In this study, Global Average pooling was adopted as the final pooling layer, wherein the pool size matches the input size, and the average value from the entire pool is selected as the output. Consequently, the number of features in the feature map undergoes significant down-sampling [73]. The feature maps' dimensions are reduced by using pooling layers. As a result, the number of parameters to learn and the amount of processing in the network are reduced [74]. All neurons are interconnected in fully connected layers. It accepts flattened output from the convolutional layers. Also, it oversees summarizing the neural network's learned characteristics. The number of parameters in the fully linked layer is computed as follows, assuming the lengths of the input and output vectors are *M* and *N*, respectively:

$$Q = M \times N + N$$

(13)

In a neural network, the *activation function* decides whether the incoming signals have reached the threshold and should output signals for the next level [75]. An artificial neural network's activation function is a function that is introduced to assist the network in learning complicated patterns in the input. Compared to a neuron-based model seen in human brains, the activation function is responsible for choosing what is to be fired to the next neuron after the process [76]. The next layer is *Dropout* which is a method for dealing with both of the following problems. It eliminates overfitting and allows for the efficient combination of exponentially many distinct neural network designs. The word *dropout* refers to units in a neural network that are no longer active (both hidden and apparent) [77]. After that, the learning process of an LSTM neural network may learn complicated associations between features and tags, but it is susceptible to time steps, the number of hidden layers, and the number of nodes in each hidden layer. The last layer is the *pooling* layer, where the entire pooled feature map matrix is flattened into a single column, which is then given to the neural network for processing.

*3.5. Classification*

Following classifiers have been chosen on the basis of published study *[30]*, which compared 179 classifiers and their performances.

### 3.5.1. Random Forest (RF)

This is an ensemble technique, which is essentially a collection of multiple decision tree algorithms that can detect on its own [78]. To discover a standard answer, the outputs of all decision trees are combined. This approach provides an additional degree of unpredictability through the involvement of predictor values set for each decision tree. Node splitting in decision trees is dependent on the best among the existing variables; however, in random forests [79], the best among a collection of randomly chosen predictors at each node is used. All characteristics must be examined for split selection during the tree-building process by picking a particular decision tree. On the other hand, Random Forest employs a subset of characteristics extracted from random decision trees. The connection between trees gives an error for the classifier in the RF method. Whereas a random forest comprises many decision trees, it contains a vast number of potential characteristics derived from various nodes. The random forest classifier is a robust method that outperforms multiple other classifiers in classification accuracy. This classifier reveals advanced precision by comparing it to others. The following parameters have been used: estimators=300, maximum depth=100, and minimum sample split=3 [80]. Below are the procedures for Random Forest categorization [81].

1. Choose $K$ features at random from a total of m features. Where $k << m$.
2. Using the most excellent rift point, calculate the node d among the $K$ features.
3. Split the node into daughter nodes using the most excellent rift.
4. Repeat steps 1–3 until the $l$ number of nodes is attained.
5. Establish a forest by repeating steps 1–4 for a $n$ number of times, resulting in a $n$ number of trees.

The following are the steps in the random forest prediction process [81]:

1. Takes the test characteristics and predicts the outcome using the rules of each randomly generated decision tree, then saves the predicted result (target).
2. Evaluate the votes for each potential target.
3. Take the random forest algorithm's final forecast as the highest chosen expected objective.

### 3.5.2. Support Vector Machine (SVM)

The basic idea of SVM is to use a decision surface to produce a hyperplane for classification, where categorization is accomplished by maximizing the isolation among various hyperplane categories [82], [83]. This classifier starts by translating the input data to a new space with more dimensions than the input. Support vectors are data points that are closest to the separating hyperplane and sit on the slab's border. The primary aim of SVM is to find the best hyperplane f(x) that separates the two input data classes [84]. SVM is a supervised machine learning approach that finds the best separating hyperplane for classifying data by maximizing the margin between classes in the feature space [85]. Hyperplane given by SVM can classify classes linearly and non-linearly. The convergence criterion in SVM was based on the following equation [88]:

$$\Delta = \frac{J(\beta) + L(\alpha)}{J(\beta) + 1} \tag{14}$$

J(β): The primal objective, L(α): Dual objective solved by Lagrangian function, and Δ: The feasibility gap.

### 3.6. Ensemble Method

Model stacking is a technique for enhancing model predictions that involves merging the outputs of numerous models and putting them through a meta-learner, which is a machine learning model. Stacking models function by running the output of several models through a "meta-learner" (usually a linear regressor/classifier, but other models such as decision trees can also be employed). The meta-learner minimizes all individual model weaknesses while maximizing their strengths. In most circumstances, the result is a robust model that generalizes well to new data *[89]*. Figure 6 shows the stack model architecture.

Different weak learners are fitted independently from each other in stacking methods, and a meta-model is trained on top of that to predict outputs based on the outputs returned by the base models. Fitting a stacking ensemble of *N* weak learners requires following steps:

• Split the training data in two folds. (*test size=0.25*)

• Select *N* weak learners and fit them to first-fold data. *(Sequential model with four dense layers and activation functions of relu and softmax. Later complied with categorical cross entropy loss and adam optimizer. Fitting the model for 50 epochs, zero verbose=0, and 1000 steps per epoch.)*

• Make predictions for observations in the second fold for each of the *N* weak learners.

• fit the meta-model (stacked model) on the second fold, using weak learner predictions as inputs.

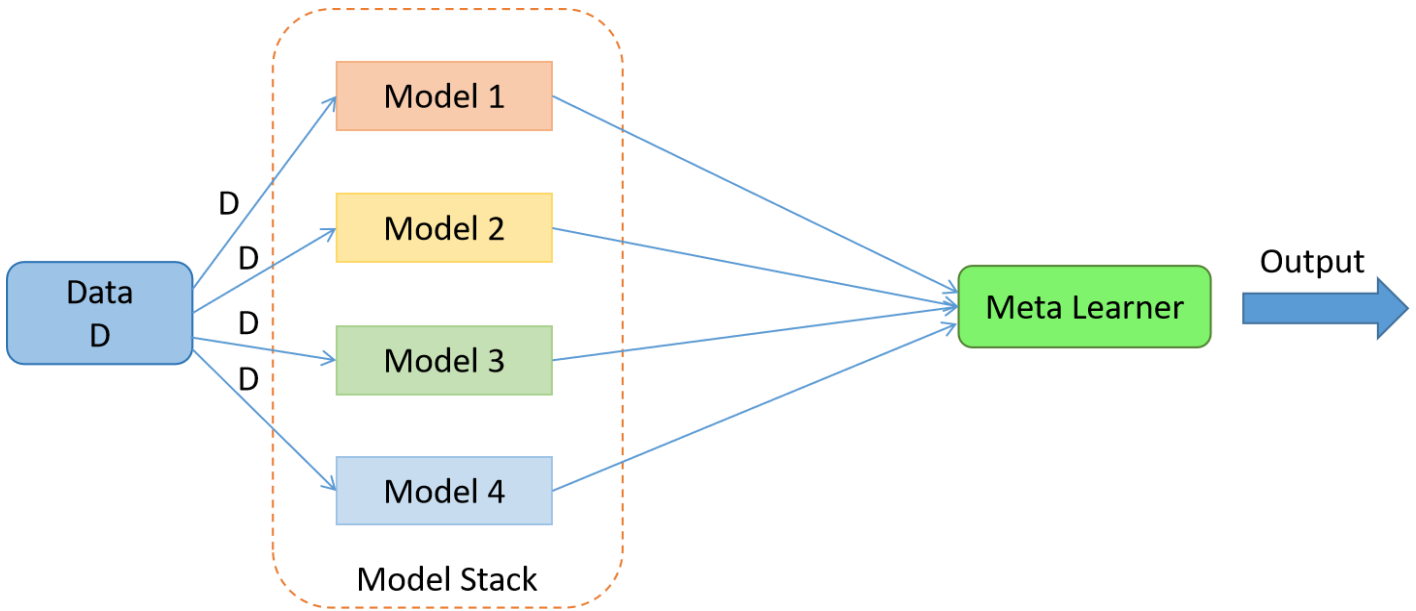

**Figure 6**. The architecture of a Stacked Model *[89]*.

### 3.7. Performance Measurements

The functionality of a classification model (or "classifier") on a set of test data, for which the absolute values are well-known, is known as a confusion matrix [90]. The following parameters have been calculated using a confusion matrix and tabulated in Table 8.

#### 3.7.1. Precision

It is the ratio of correctly predicted positive observations to total anticipated positive observations. The low False-Positive (FP) rate is related to the high accuracy. The following formula shows how to calculate it.

$$Precision = \frac{True\ Positive}{True\ Positive + False\ Positive}$$

(15)

#### 3.7.2. Recall (Sensitivity)

The ratio of correctly predicted positive observations to all observations in the actual class.

$$Recall = \frac{True\ Positive}{True\ Positive + False\ Negative}$$

(16)

#### 3.7.3. F1-Score

The weighted average of Precision and Recall is used to get the F1-Score. Thus, both false positives and false negatives are included while calculating this score.

$$F1\ Score = 2\frac{(Recall \times Precision)}{(Recall + Precision)}$$

(17)

#### 3.7.4. Specificity

Specificity is the correct negative labeled by the program. Also, it is a probability of a proper test result in subjects without a condition.

$$Specificity = \frac{True\ Negative}{True\ Negative + False\ Positive}$$

(18)

### 3.7.5. Accuracy

The most straightforward intuitive performance metric is accuracy, which is just the ratio of correctly predicted observations to all observations.

$$Accuracy = \frac{True\ Negative + True\ Positive}{Total}$$

(19)

## 4. Results

In this section, pre-processing and NN modeling results have been discussed. Simulations have been conducted using Jupiter Notebook and Spyder for Python programming via Anaconda Navigator.

### 4.1. Pre-processing Results

As discussed earlier, we had three pre-processing steps for raw PPG signals. This process prepares the signal for the deep learning process.

As depicted in Figure 7, the dataset has raw PPG signal with noise, which will be passed through the median filter, later through low pass Chebyshev filter, and then applying backward forward filter.

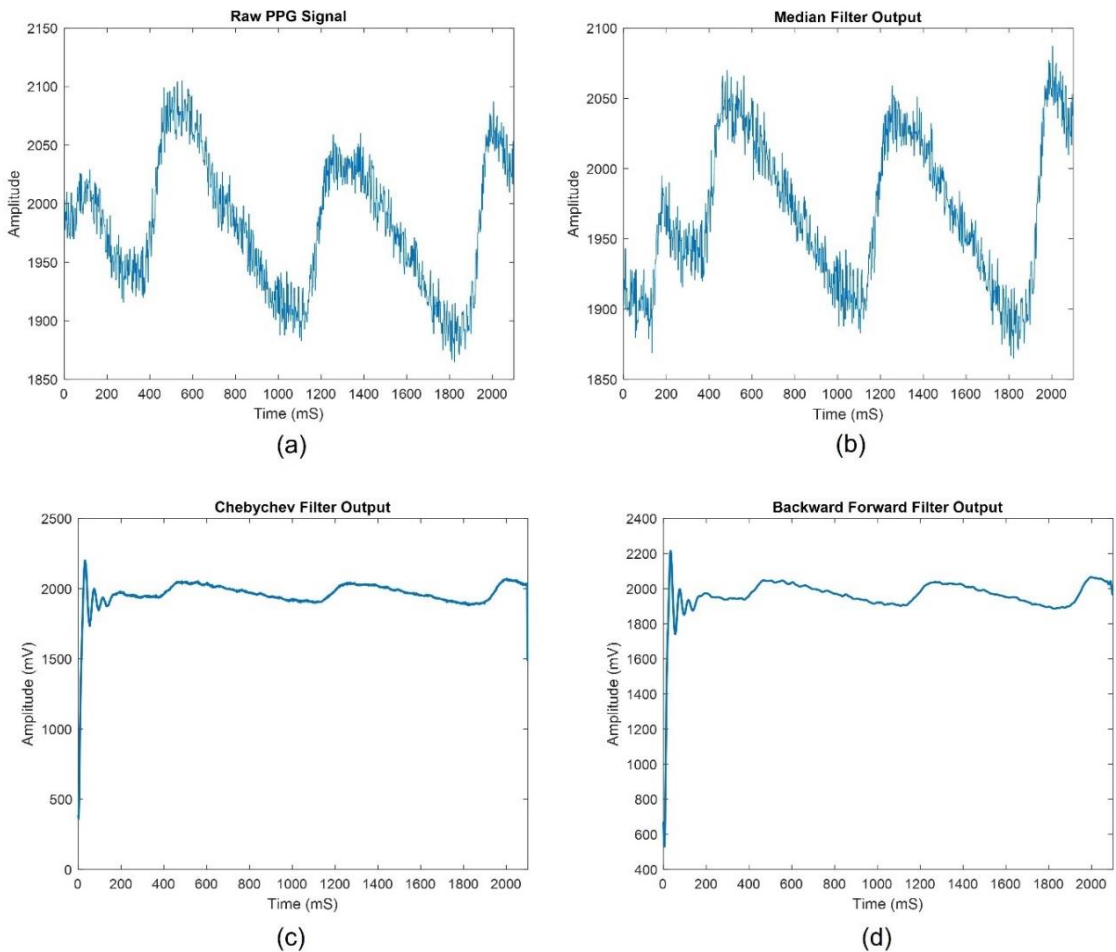

**Figure 7.** Depiction of a) PPG signal with noise, b) PPG signal after applying a median filter, likewise applying c) low pass Chebyshev filter, and then d) applying backward forward filter.

## 4.2. Neural Network Layers

A neural network is composed of three types of layers: the input layer that store the neural network's initial data, the output layer, which produces a result for given inputs; and hidden layers which lies between the input and output layers, where all the processing takes place [91]. Table S1 to S5 in Supporting file shows neural network layers for CNN, LSTM, BiLSTM, STFT with CNN combination, and LSTM with CNN combination, including both output shape and number of parameters involved. Various layers have been used, equaling 12, 5, 12, 9, and 11. Furthermore, Table 3 provides the performance measurements for all neural networks by tabulating crucial metrics, such as precision, recall, f-score, specificity, and accuracy. Two batch sizes have been used to check the variation of paraments. Batch size states the number of samples processed before a model is updated. Batch sizes of 3 and 16 have been examined for all techniques.

## 4.3. Ensemble Stacking Output

Table 3 also shows the results of the stacked model for different classifiers. All the meta models performed better than standard models especially LSTM, LSTM-CNN-RF, STFT-CNN-SVM.

**Table 3.** PERFORMANCE MEASUREMENTS FOR ALL NEURAL NETWORKS.

| | CNN | | | | CNN + SVM | | | | CNN+ RF | | | |
|---|---|---|---|---|---|---|---|---|---|---|---|---|
| | Pre-Hypertension | | Hypertension | | Pre-Hypertension | | Hypertension | | Pre-Hypertension | | Hypertension | |
| **Batch Size** | **16** | **3** | **16** | **3** | **16** | **3** | **16** | **3** | **16** | **3** | **16** | **3** |
| **Precision (%)** | 63.4 | 59.6 | 56.8 | 70.6 | 63.0 | 67.8 | 56.8 | 61.7 | 63.0 | 66.7 | 61.0 | 70.6 |
| **Recall (%)** | 53.6 | 81.0 | 45.5 | 43.6 | 69.0 | 47.6 | 45.5 | 52.7 | 69.0 | 57.1 | 45.5 | 43.6 |
| **F Score (%)** | 58.1 | 68.7 | 50.5 | 53.9 | 65.9 | 55.9 | 50.5 | 56.8 | 65.9 | 61.5 | 52.1 | 53.9 |
| **Specificity (%)** | 60.0 | 29.2 | 71.2 | 84.8 | 47.7 | 70.8 | 71.2 | 72.7 | 47.7 | 63.1 | 75.8 | 84.8 |
| **Accuracy (%)** | 56.4 | 58.4 | 59.5 | 66.1 | 59.7 | 57.7 | 55.4 | 63.6 | 59.7 | 59.7 | 62.0 | 66.1 |
| | **LSTM** | | | | **LSTM + SVM** | | | | **LSTM + RF** | | | |
| | Pre-Hypertension | | Hypertension | | Pre-Hypertension | | Hypertension | | Pre-Hypertension | | Hypertension | |
| **Batch Size** | **16** | **3** | **16** | **3** | **16** | **3** | **16** | **3** | **16** | **3** | **16** | **3** |
| **Precision (%)** | 63.3 | 75.4 | **100.0** | 61.9 | 61.2 | 69.2 | **100.0** | **100.0** | 62.8 | 70.3 | 60.6 | 47.6 |
| **Recall (%)** | **82.1** | 58.3 | 1.8 | 23.6 | 97.6 | 64.3 | 1.8 | 10.9 | 64.3 | 61.9 | 36.4 | 36.4 |
| **F Score (%)** | 71.5 | 65.8 | 3.5 | 34.2 | 75.2 | 66.7 | 3.5 | 19.7 | 63.5 | 65.8 | 45.5 | 41.3 |
| **Specificity (%)** | 38.5 | 75.4 | **100.0** | 87.9 | 20.0 | 63.1 | **100.0** | **100.0** | 50.8 | 66.2 | 80.3 | 66.7 |
| **Accuracy (%)** | 63.1 | 65.8 | 55.4 | 58.7 | 63.8 | 63.8 | 55.4 | 59.5 | 58.4 | 63.8 | 60.3 | 52.9 |
| | **LSTM-CNN** | | | | **LSTM-CNN + SVM** | | | | **LSTM-CNN + RF** | | | |
| | Pre-Hypertension | | Hypertension | | Pre-Hypertension | | Hypertension | | Pre-Hypertension | | Hypertension | |
| **Batch Size** | **16** | **3** | **16** | **3** | **16** | **3** | **16** | **3** | **16** | **3** | **16** | **3** |
| **Precision (%)** | 68.5 | 68.9 | 64.2 | 73.2 | 65.2 | 71.0 | 63.5 | 73.3 | 61.1 | 71.0 | 66.1 | 72.3 |
| **Recall (%)** | 44.0 | 60.7 | 61.8 | 54.5 | 51.2 | 58.3 | 60.0 | 60.0 | 52.4 | 58.3 | 67.3 | 61.8 |
| **F Score (%)** | 53.6 | 64.5 | 63.0 | 62.5 | 57.4 | 64.0 | 61.7 | 66.0 | 56.4 | 64.0 | 66.7 | 66.6 |
| **Specificity (%)** | 73.8 | 64.6 | 71.2 | 83.3 | 64.6 | 69.2 | 71.2 | 81.8 | 56.9 | 69.2 | 71.2 | 80.3 |
| **Accuracy (%)** | 57.0 | 62.4 | 66.9 | 70.2 | 57.0 | 63.1 | 66.1 | **71.9** | 54.4 | 63.1 | 69.4 | **71.9** |
| | **Bi LSTM** | | | | **Bi LSTM + SVM** | | | | **Bi LSTM + RF** | | | |
| | Pre-Hypertension | | Hypertension | | Pre-Hypertension | | Hypertension | | Pre-Hypertension | | Hypertension | |
| **Batch Size** | **16** | **3** | **16** | **3** | **16** | **3** | **16** | **3** | **16** | **3** | **16** | **3** |
| **Precision (%)** | 59.8 | 71.6 | 85.7 | 64.8 | 58.0 | 75.0 | 55.6 | 63.6 | 60.9 | 74.2 | 57.5 | 69.2 |
| **Recall (%)** | 72.6 | 57.1 | 48.3 | 63.6 | 77.4 | 57.1 | 9.1 | 63.6 | 63.1 | 54.8 | 41.8 | 65.5 |
| **F Score (%)** | 65.6 | 63.5 | 61.8 | 64.2 | 66.3 | 64.8 | 15.6 | 63.6 | 62.0 | 63.0 | 48.4 | 67.3 |
| **Specificity (%)** | 36.9 | 70.8 | 89.4 | 71.2 | 27.7 | 75.4 | 93.9 | 69.7 | 47.7 | 75.4 | 74.2 | 75.8 |
| **Accuracy (%)** | 57.0 | 63.1 | 66.0 | 67.8 | 55.7 | 65.1 | 55.4 | 66.9 | 56.4 | 63.8 | 59.5 | 71.1 |
| | **STFT-CNN** | | | | **STFT-CNN +SVM** | | | | **STFT-CNN + RF** | | | |

| | Pre-Hypertension | | Hypertension | | Pre-Hypertension | | Hypertension | | Pre-Hypertension | | Hypertension | |
|---|---|---|---|---|---|---|---|---|---|---|---|---|
| **Batch Size** | **16** | **3** | **16** | **3** | **16** | **3** | **16** | **3** | **16** | **3** | **16** | **3** |
| **Precision (%)** | 62.6 | 64.1 | 55.4 | 52.2 | 61.5 | 58.1 | 54.7 | 47.3 | 61.1 | 58.0 | 55.6 | 52.2 |
| **Recall (%)** | 77.0 | 67.6 | 67.4 | 52.2 | 79.7 | 67.6 | 63.0 | 56.5 | 74.3 | 68.9 | 65.2 | 52.2 |
| **F Score (%)** | 69.1 | 65.8 | 60.8 | 52.2 | **69.4** | 62.5 | 58.6 | 51.5 | 67.1 | 63.0 | 60.0 | 52.2 |
| **Specificity (%)** | 54.7 | 62.7 | 66.7 | 70.7 | 50.7 | 52.0 | 68.0 | 61.3 | 53.3 | 50.7 | 68.0 | 70.7 |
| **Accuracy(%)** | 65.8 | 65.1 | 66.9 | 63.6 | 65.1 | 59.7 | 66.1 | 59.5 | 63.8 | 59.7 | 66.9 | 63.6 |

**Table 3.** PERFORMANCE MEASUREMENTS FOR ALL NEURAL NETWORKS

| ENSEMBLE MODELS | Accuracy | Precision | Recall | F1 score |
|---|---|---|---|---|
| **Meta CNN** | 0.9875 | 0.9834 | 0.9875 | 0.9846 |
| Meta CNN-SVM | 0.9893 | 0.9829 | 0.9819 | 0.9832 |
| Meta CNN-RF | 0.9923 | 0.995 | 0.991 | 0.9899 |
| **Meta LSTM** | 0.9661 | 0.9723 | 0.9678 | 0.9748 |
| Meta LSTM-SVM | 0.9812 | 0.9851 | 0.9821 | 0.9820 |
| Meta LSTM-RF | 1.0 | 1.0 | 1.0 | 1.0 |
| **Meta LSTM-CNN** | 0.9897 | 0.9828 | 0.9972 | 0.9934 |
| Meta LSTM-CNN-SVM | 0.9998 | 0.9964 | 0.9971 | 0.9890 |
| Meta LSTM-CNN-RF | 1.0 | 1.0 | 1.0 | 1.0 |
| **Meta BiLSTM** | 0.9160 | 0.913 | 0.9182 | 0.9138 |
| Meta BiLSTM-SVM | 0.9312 | 0.9562 | 0.9416 | 0.9206 |
| Meta BiLSTM-RF | 0.9316 | 0.9351 | 0.9289 | 0.9379 |
| **Meta STFT-CNN** | 0.9752 | 0.9840 | 0.9873 | 0.9711 |
| Meta STFT-CNN-SVM | 1.0 | 1.0 | 1.0 | 1.0 |
| Meta STFT-CNN-RF | 0.9867 | 0.9823 | 0.9974 | 0.9912 |

## 5. Discussion

Researchers [84] devised and implemented various methods for predicting systolic and diastolic blood pressure using PPG signal features and machine learning techniques. This effectively illustrates how the PPG signal may be utilized for noninvasively estimating blood pressure without using cuff-based pressure measurement. Such systems can aid in the constant monitoring of blood pressure and avoid any severe health problems caused by abrupt fluctuations [92]. The proposed models were trained in 100 epochs for the 16- batch size and 300 epochs for the three-batch size. Overfitting occurs when the model has high training accuracy and low training loss. To avoid such a problem, data was normalized before training, dropout layer and an early stoppage criterion function were used. Table 8 provides the performance measurements for all neural networks by calculating precision, recall, f-score, specificity, and accuracy. Results showed that LSTM-CNN+RF and Bi LSTM+RF performed the best, achieving 71.9% and 71.1%, respectively. The dataset is medium-sized, which somehow influences the performance. It impacts the quality of the mapping function approximated by neural networks, along with the quality of the estimated performance of a fit neural network model.

### 5.1. Comparison With Other Studies

Table 10 compares the proposed work and other studies using the same dataset, Guilin People's Hospital in Guilin, China, with 219 subjects [55]. The first study by Martinez-Rios et al. [93] used Wavelet Scattering Transform (WST) as a feature extraction from PPG signals and other clinical variables such as age, weight, and heart rate to train a single SVM model. The trained model was limited to classifying two classes only, normotension and pre-hypertension; therefore, it can classify patients with hypertension. Overall, the results achieved are almost the same or lower than ours except for the precision, which is expected in a minor classification problem. The second study by Yen et al. [94] tried different deep learning models and parameters to classify hypertension into four levels. Maximum accuracy of 76% was archived by using Xception + BILSTM with 37 layers and 200 epochs. Compared to our model, we have a simpler model consisting of three layers with 100 epochs only, which indicates a faster simulation time. Zhang et al. [95] used wavelet transform (WT) for signal pre-processing then LSTM and BILSTM for three-level classification, achieving an accuracy of 68.83% and 69.5%, respectively. A detailed comparison of various BP and PPG studies can be found in [96] which discusses results in absolute error.

**Table 10.** COMPARISON WITH OTHER RECENT STUDIES OF THE SAME DATASET.

| Ref | Classifier Used | Accuracy | Precision | Recall | F1-Score |
|---|---|---|---|---|---|
| [93] | SVM | 71.42% | 84.61% | 52.38% | 64.70% |
| [94] | BILSTM | 76% | 48% | 45% | NA |
| [95] | LSTM and BILSTM | 68.83% and 69.58% | 74.21% and 72.46% | 78.08% and 81.84% | NA |
| **This work** | LSTM-CNN+SVM | 71.9% | 73.3% | 60.0% | 66.0% |

*5.2. Tradeoffs Of Performance Metrics*

Precision can be adjusted by changing the model's parameters and hyperparameters. While adjusting, it can be noticed that a greater precision generally results in a lower recall, and a higher recall, in turn, results in lower accuracy. Similarly, the recall value of any machine learning model also can be altered by adjusting multiple parameters or hyperparameters. A higher or lower recall has a specific meaning for any model: A high recall indicates that most positive instances (TP+FN) will be identified as such (TP). This results in a more significant number of FP measurements and a decrease in overall accuracy. However, Suppose the result is a low recall. In that case, it signifies that many FNs (should have been positive but labeled negative), which means that if results find a positive example, it may be more confident that it is a real positive.

Furthermore, F1 is not as intuitive as accuracy, but it's typically more beneficial, especially if the class distribution is unequal. When false positives and false negatives cost the same, accuracy is maximized. Therefore, it is best to look at both Precision and Recall if the cost of false positives and false negatives are substantially different. Additionally, sensitivity (recall) and specificity are inversely related. Susceptible tests get more positive outcomes for patients with illness, although precise tests shows patients without a finding having no illness. Therefore, sensitivity and specificity should always be considered simultaneously to provide a complete diagnosis. Besides, accuracy is a good quality measure when datasets are symmetric where values of false-positive and false-negatives are nearly similar. Hence, other parameters are also crucial to evaluating the performance of a model.

*5.3. Pros And Cons of Neural Network Used*

Deep CNNs' popularity and wide application fields can be accredited to the benefit mentioned in the following points: CNNs combine feature extraction and classification operations into a single learning body. They could optimize the characteristics straight from the raw input during the training phase. In contrast to fully connected Multi-Layer Perceptron's (MLP) networks, CNN neurons are sparsely linked with coupled weights, allowing them to analyze massive inputs with significant computing efficiency. CNNs are unaffected by small data changes, for instance, skewing, scaling, translation, and distortion. CNNs can adjust to a variety of input sizes. Nevertheless, the significant benefit of CNN over its predecessors is that it recognizes significant characteristics without the need for human intervention. In critical signal processing applications like patient-specific PPG categorization, fundamental health monitoring, detecting anomalies in power electronics circuits, and motor-fault detection, 1D Convolutional Neural Networks (CNNs) have become advanced technology. Furthermore, 1D CNNs may be employed straight to the raw signals (e.g., current, voltage, vibration, etc.), i.e., implementing the process without making any pre- or post-processing, for example, feature selection, dimension reduction, extraction, denoising, etc. [97]. Likewise, a real-time and inexpensive hardware implementation is possible because of the simplicity and efficient setup of such adaptive 1D CNNs that execute just linear 1D convolutions (i.e., additions & scalar multiplications) [98]. LSTM can make an instant prediction as all data available are taken from the past. Therefore, it only predicts future data. Nevertheless, it takes longer to train, as it needs additional RAM; moreover, it is easy to overfit. However, BiLSTM solves the fixed sequence to sequence prediction issue as it can predict the past data from future data and vice versa. The size of both the input and output in a vanilla RNN is limited. On the other hand, Bi-LSTM is expensive since it is more complex and needs substantial computational power to be trained than LSTM as it has two LSTM cells. When it comes to rendering power, the STFT has an advantage. Integration of a 3D wavelet spectrum to obtain power is conceivable, just as the integration of the STFT to obtain power information from the volume under the surface is possible. However, due to the multiresolution nature of wavelet analysis, calculating relative powers directly from the size of the 3D peaks in the wavelet spectrum is challenging. The power is directly proportional to the height of the peaks, which is a property of the STFT.

*5.4. Ensemble Stacking*

Stacked generalization, commonly known as stacking, is a machine learning ensemble algorithm. Using a meta-learning method, it learns how to aggregate predictions from two or more underlying ML systems. The benefit of stacking is that it can combine the capabilities of numerous high-performing models to create predictions on a classification or regression job that surpass any single model in the ensemble. Model performance can be easily extracted using stacking models. In some data science applications where all types of performance important, stacking models can be a simple and convenient technique to achieve this. Stacking models, on the other hand, often take longer to train and have substantially slower latencies than other models.

*5.5. Hardware Implementation: Future Scope*

The requirement for high-performance processing resources is one of the significant roadblocks to AI's full potential. Hardware accelerators have recently been created to supply the computing capacity required by AI and ML technologies. FPGAs, GPUs, and ASICs create hardware accelerators that speed up computationally heavy processes. These accelerators offer high-performance hardware while yet maintaining the needed precision. An optimized and customized hardware implementation can lower system costs by reducing the needed resources and lowering power consumption (while enhancing performance) [99]. TensorFlow, Keras, and Caffe are examples of development platforms used to create and test AI algorithms. Some of these platforms are Visual Geometry Group's AlexNet and VGG tools, popular standard neural networks (NN). Hardware accelerators are created when developed NNs for a specific range of applications are implemented in hardware. FPGAs, GPUs, and ASICs are being explored for hardware implementations, each with its own set of advantages and disadvantages. Compared to ordinary CPUs, which may be 1/10 1/100 times slower [98], these hardware accelerators use parallelism to improve throughput and give significantly better performance. In particular, deep learning techniques need much storage to leverage big datasets and intense computer processing to create autonomous adaptive, innovative systems with extraordinarily accurate and human-like behavior [100]. It is worth noting that deep learning is a very new field [101]. Various parameters are described in following Table 8, related to the hardware perspective.

# 6. Conclusion

This study presents computationally efficient neural network models (Convolution Neural Network, Long Short-Term Memory, Bidirectional Long Short-Term Memory) for classifying hypertension (Stage I and Stage II) patients. It illustrates how the 1D time-series data (PPG) signal can detect blood pressure using various machines and deep learning techniques. Simulations have been done on the dataset with the combinations, two batch sizes (16 and 3), two machine learning classifiers (Random Forest and Support Vector machine learning), along with NNs and Short-time Fourier transform (STFT). The LSTM model provides the best results among all combinations of Neural Networks, with precision and specificity of 100 percent and recall of 82.1 percent. On the other hand, the LSTM-CNN model achieves a maximum accuracy of 71.9 percent. This study can benefit researchers in healthcare and machine learning, especially with signal analysis. Deep learning is trial and error method, which can be modified and improved in endless ways. Moreover, for the limitations of the study the dataset is normal size, this can be improved using big data. Besides complexity can be a limitation for hardware implementation, nevertheless, quantization approach on deep neural networks reduces the precision of the weights, biases, and activations such that they consume less memory. Hence converts complex deep learning models easy feedable for microcontrollers [102]. For the future scope of the study, hardware implementation has been explored (i.e., ASIC, FPGA, GPU), which can give this study a new dimension. In addition, metadata of dataset can be used to provide further insights in this study.

**Data Availability**: This data is publicly available on https://www.nature.com/articles/sdata201820